# Tree-dependent Component Analysis

**Francis R. Bach**
Computer Science Division
University of California
Berkeley, CA 94720, USA
fbach@cs.berkeley.edu

**Michael I. Jordan**
Computer Science Division
and Department of Statistics
University of California
Berkeley, CA 94720, USA
jordan@cs.berkeley.edu

## Abstract

We present a generalization of independent component analysis (ICA), where instead of looking for a linear transform that makes the data components independent, we look for a transform that makes the data components well fit by a tree-structured graphical model. Treating the problem as a semiparametric statistical problem, we show that the optimal transform is found by minimizing a contrast function based on mutual information, a function that directly extends the contrast function used for classical ICA. We provide two approximations of this contrast function, one using kernel density estimation, and another using kernel generalized variance. This *tree-dependent component analysis* framework leads naturally to an efficient general multivariate density estimation technique where only bivariate density estimation needs to be performed.

## 1 INTRODUCTION

Given a multivariate random variable $x$ in $\mathbb{R}^m$, independent component analysis (ICA) consists in finding a linear transform $W$ such that the resulting components of $s = Wx = (s_1, \ldots, s_m)^T$ are as independent as possible. It has been applied successfully to many problems where it can be assumed that the data are actually generated as linear mixtures of independent components, such as audio blind source separation or biomedical imagery (see e.g. Hyvärinen et al., 2001). It can also be used as a general multivariate density estimation method where, once the optimal transformation $W$ has been found, only univariate density estimation needs to be performed. In this paper, we generalize these ideas: we search for a linear transform $W$ such that the components of $s = Wx = (s_1, \ldots, s_m)^T$ can be well modelled by a tree-structured graphical model. We don't fix the topology of the tree in advance; rather, we search for the best possible tree in a manner analogous to the Chow-Liu algorithm (Chow and Liu, 1968), which indeed serves as an inner loop in our algorithm. We refer to this methodology as *tree-dependent component analysis (TCA)*.

By removing the strong assumption made by ICA that the underlying components are independent, TCA can be applied to a wider range of problems in which the data are transformed by an unknown linear transformation. For example, in a given musical piece, instruments are not mutually independent. Modelling their dependencies is necessary to achieve successful demixing, and the TCA model provides a principled approach to deal with such dependencies. Alternatively, as with ICA, the TCA approach can be used as an efficient method for general multivariate density estimation. Indeed, once the linear transform $W$ and the tree $T$ are found, we need only perform *bivariate* density estimation, skirting the curse of dimensionality while obtaining a flexible model. The models that we obtain using these two stages—first find $W$ and $T$, then estimate densities—are fully tractable for learning and inference.

While in most applications of graphical models, specific parametric distributional assumptions are made for each of the random variables in the model, we prefer to treat TCA as a *semiparametric model* (Bickel et al., 1998), in which the actual marginal and conditional distributions of the tree-dependent components are left unspecified. In the simpler case of ICA, which as a graphical model is a bipartite directed graph from nodes representing $s$ to nodes representing $x$, it is known that maximizing the semiparametric likelihood is equivalent to minimizing the mutual information between the estimated components (Cardoso, 1999). In Section 2, we review the relevant ICA results and we extend this approach to the estimation of $W$ and $T$ in the TCA model, deriving an expression for the semi-parametric likelihood which involves a number of pairwise mutual information terms corresponding to the cliques in the tree. As in ICA, to obtain a criterion that can be used to estimate the parameters in the model from data (a "contrast function"), we approximate this population likelihood. In particular, in this paper, we derive two contrast functions. In Section 4, we use kernel density estimation as plug-in estimates of the necessary mutual information terms. In Section 5, we show how the "kernel generalized variance" proposed in our earlier work on ICA (Bach and Jordan, 2001) can be extended to approximate the TCA semiparametric likelihood.

Once the contrast functions are defined, we are faced with a minimization problem with respect to $W$ and $T$. We use an alternating minimization procedure, where we alternate between minimizing with respect to $T$, essentially by the Chow-Liu algorithm, and minimizing with respect to $W$, by gradient descent. The algorithm is presented in Section 6.

The TCA model has interesting properties that differ from the classical ICA model. First, in the Gaussian case, whereas the ICA model reduces to simply finding uncorrelated components (with a lack of identifiability for specific directions in the manifold of uncorrelated components), in the TCA case there are additional solutions beyond uncorrelated components. Second, in the general non-Gaussian case, additional identifiability issues arise. We study these issues in Section 3. In Section 7, we show how, once the optimal linear transform $W$ has been obtained, conditional Gaussian mixture models can be applied to estimate the density. Finally, in Section 8, we illustrate our algorithm with simulations on synthetic examples, some of which do, and some of which do not, follow the TCA model.

## 2 SEMIPARAMETRIC MAXIMUM LIKELIHOOD

In this section we derive the objective function that will be minimized to determine the demixing matrix $W$ and the tree $T$ in the TCA model. Let $x = (x_1, \ldots, x_m)^T$ be an $m$-component random vector with joint "target" distribution $p(x)$. Our primary goal is to minimize the Kullback-Leibler (KL) divergence $D(p||q) \triangleq E_{p(x)} \log \frac{p(x)}{q(x)}$ between $p(x)$ and our model $q(x)$ of this vector. Typically, $p(x)$ will be the empirical distribution associated with a training set and minimizing the KL divergence is well known to be equivalent to maximizing the likelihood of the data. In a semiparametric model, the parameters of interest—the matrix $W$ and the tree $T$ in our case—do not completely specify the distribution $q(x)$; the additional (infinite-dimensional) set of parameters that would be needed to complete the specification are left unspecified.[1] More precisely, we define our objective function for $T$ and $W$ to be a "profile likelihood"—the minimum of the KL divergence $D(p||q)$ with respect to the distributions of the source components $s_i$. As we will show, it turns out that this criterion can be expressed in term of mutual information terms relating components that are neighbors in the tree $T$. We first review the classical ICA setting where the components $s_i$ are assumed independent. Then, we describe the case where $W$ is fixed to identity and $T$ can vary—this is simply the tree model presented by Chow and Liu (1968). We finally show how the two models can be combined and generalized to the full TCA model where both $W$ and $T$ can vary.

In the following sections $p(x_u, x_v)$ and $p(x_u)$ will denote the marginalizations of $p(x)$ on $(x_u, x_v)$ and $x_u$, and $d_u(T)$ will denote the degree of node $u$ in $T$. We will also work with the pairwise mutual information $I(x_u, x_v)$ between two variables $x_u$ and $x_v$, defined as $I(x_u, x_v) = D(p(x_u, x_v)||p(x_u)p(x_v))$ and the $m$-fold mutual information $I(x_1, \ldots, x_m)$, defined as $I(x_1, \ldots, x_m) = D(p(x)||p(x_1) \cdots p(x_m))$.

### 2.1 ICA MODEL

The classical ICA model takes the form $x = As$ where $A$ is an invertible mixing matrix and $s$ has independent components. Note that if the variable $x$ is Gaussian, then ICA reduces to principal component analysis (PCA), and the model is not identifiable, that is, the optimal matrix $W = A^{-1}$ is only defined up to a rotation matrix. Thus, non-Gaussianity of components is a crucial assumption for a full ICA solution to be well-defined, and we also make this assumption throughout the current paper. However, the Gaussian case is still of interest because it allows us to reduce the size of the search space for $W$ (see Section 3.2 and Section 9 for details).

Given a random vector $x$ with distribution $p(x)$ (not necessarily having independent components), the distribution $q(x)$ with independent components that is closest to $p(x)$ in KL divergence, is the product $q(x) = p(x_1) \cdots p(x_m)$ of the marginals of $x$ and the minimum KL divergence is thus $D(p(x)||p(x_1) \cdots p(x_n))$, which is exactly the mutual information $I(x_1, \ldots, x_m)$.

We now turn to the situation where $A$ can vary. Letting $W = A^{-1}$, we let $\mathcal{D}^W$ denote the set of all distributions $q(x)$ such that $s = Wx$ has independent components. Since the KL divergence is invariant by an invertible transformation, the best approximation to $p(x)$ by a distribution in $\mathcal{D}^W$ is obtained as the

---

[1] Note that the focus is on general *continuous* random variables in the ICA and TCA settings.

product of the marginals of $s = Wx$, which yields:

$$\min_{q \in \mathcal{D}^W} D(p||q) = I(s_1, \ldots, s_m). \quad (1)$$

Thus, in the semiparametric ICA approach, we wish to minimize the mutual information of the estimated components $s = Wx$. We will generalize Eq. (1) to the TCA model in Section 2.3.

In practice, we do not know the density $p(x)$ and thus the estimation criteria must be replaced by functionals of the sample data, functionals that are referred to as "empirical contrast functions". Classical ICA contrast functions involve either approximations to the mutual information or alternative measures of dependence involving higher-order moments (Hyvärinen et al., 2001, Cardoso, 1999).

## 2.2 CHOW-LIU MODEL

Given an undirected spanning tree $T$ on the vertices $\{1, \ldots, m\}$, we let $\mathcal{D}^T$ denote the set of probability distributions $q(x)$ that *factorize* in $T$; i.e., the set of distributions that can be written in the form $q(x) \propto \prod_{(u,v) \in T} \varphi_{uv}(x_u, x_v)$, where the potentials $\varphi_{uv}$ are arbitrary functions. We want to model $p(x)$ using a distribution $q(x)$ in $\mathcal{D}^T$. Trees are a special case of *decomposable models* and thus, for a given tree $T$, minimizing the KL divergence is straightforward and yields the following "Pythagorean" expansion of the KL divergence (Jirousek, 1991):

**Theorem 1** *For a given tree $T$ and a target distribution $p(x)$, we have, for all distributions $q \in \mathcal{D}^T$,*

$$D(p||q) = D(p||p_T) + D(p_T||q) \quad (2)$$

*where $p_T(x) = \frac{\prod_{(u,v) \in T} p(x_u, x_v)}{\prod_u p(x_u)^{d_u(T)-1}}$. In addition, $q = p_T$ minimizes $D(p||q)$ over $q \in \mathcal{D}^T$, and we have:*

$$I^T(x) \triangleq \min_{q \in \mathcal{D}^T} D(p||q) = D(p||p_T) \quad (3)$$

$$= I(x_1, \ldots, x_m) - \sum_{(u,v) \in T} I(x_u, x_v) \quad (4)$$

We refer to $I^T(x)$ as the *T-mutual information*: it is the minimum possible loss of information when encoding the distribution $p(x)$ with a distribution that factorizes in $T$. It is equal to zero if and only if $p$ does factorize in $T$. Such a quantity can be defined for any directed or undirected graphical model (cf. Friedman et al., 2001). When the graphical model is a tree, the problem of minimizing $I^T(x)$ with respect to $T$, given the pairwise mutual informations, is a maximum weight spanning tree problem (Chow and Liu, 1968), which is readily solved in polynomial time by simple greedy algorithms (Cormen et al., 1989).

## 2.3 TCA MODEL

In TCA, we wish to model the variable $x$ using the model $x = As$, where $A$ is an invertible mixing matrix and $s$ factorizes in a tree $T$. Letting $W = A^{-1}$, we let $\mathcal{D}^{W,T}$ denote the set of all such distributions. The KL divergence is invariant by invertible transformation of its arguments, so Theorem 1 can be easily extended:

**Theorem 2** *If $x$ has distribution $p(x)$, then the minimum KL divergence between $p(x)$ and a distribution $q(x) \in \mathcal{D}^{W,T}$ is equal to the T-mutual information of $s = Wx$, that is:*

$$J(x, W, T) \triangleq \min_{q \in \mathcal{D}^{W,T}} D(p||q) = I^T(s) \quad (5)$$

$$= I(s_1, \ldots, s_m) - \sum_{(u,v) \in T} I(s_u, s_v) \quad (6)$$

Therefore, in the semiparametric TCA approach, we wish to minimimize $J(x, W, T)$ with respect to $W$ and $T$.

As in ICA, we do not know the density $p(x)$ and the estimation criteria must be replaced by empirical contrast functions. In the TCA setting, it is important that we maintain a link with mutual information: indeed the interplay between the 2-fold and $m$-fold mutual information terms is crucial, making it possible to avoid overcounting or undercounting the pairwise dependencies. The contrast functions that we propose thus have such a link—our first contrast function approximates the mutual information terms directly, and our second proposed contrast function has an indirect link to mutual information. Before describing these two contrast functions, we turn to the description of the main properties of the TCA model.

## 3 PROPERTIES OF THE TCA MODEL

In this section we describe some properties of the TCA model, relating them to properties of the simpler ICA model. In particular we focus on identifiability issues and on the Gaussian case.

### 3.1 IDENTIFIABILITY ISSUES

In the ICA model, it is well known that the matrix $W$ can only be determined up to permutation or scaling of its rows. In the TCA model, we have the following indeterminacies.

**Permutation of components.** $W$ can be premultiplied by a permutation matrix without changing the value of $J(x, W, T)$, as long as the tree $T$ is also permuted analogously. This implies that in principle we don't have to consider all possible trees, but just equivalence classes under vertex permutation.

**Scaling of the components.** $W$ can be premultiplied by any invertible diagonal matrix. Thus we can restrict our search to components that have unit variance.

**Mixing of a leaf node with its parent.** For a given tree structure $T$, and a leaf node $c$, adding a multiple of the value of its parent $p$ to the value of the leaf will not change the goodness of fit of the tree $T$. Indeed, a leaf node is only present in the likelihood through the conditional probability $p(s_c|s_p)$, and thus we can equivalently model $p(s_c|s_p)$ or $p(\lambda s_c + \mu s_p|s_p)$ for any $\mu$ and any non-zero $\lambda$. The $T$-mutual information $I^T(s)$ is thus invariant under such transformations.

While the first two identifiability issues are easily handled by simple conventions, the third indeterminacy is not easily removed via a simple convention; indeed, the empirical contrast functions that we develop do not respect the mixing invariance. We could deal with the issue by "normalizing" the relation between a leaf and its parent, for example by requiring marginal decorrelation. However, this normalization depends on the tree, so it is not appropriate when comparing trees. Instead, we simply add a penalty term to our contrast functions, penalizing the correlation between components that are linked by an edge of the tree $T$ (see Section 6 for details).

### 3.2 THE GAUSSIAN CASE

In the ICA model, if the variable $x$ is Gaussian, ICA reduces to PCA and the solutions are the matrices $W$ that make the covariance matrix of $s = Wx$ equal to identity. Thus, they are defined up to a rotation matrix $R$.

In the TCA model, with a fixed tree $T$, there is more than one covariance matrix that leads to a tree-structured graphical model with graph $T$ for the underlying Gaussian variable. For a Gaussian variable, conditional independences can be read out from zeros in the inverse of the covariance matrix (e.g. Lauritzen, 1996). Applying to trees, we get:

**Proposition 1** *If $x = (x_1, \ldots, x_m)$ is Gaussian with covariance matrix $\Sigma$, it factorizes in the tree $T$ if and only if for all $(u,v) \notin T$, we have $(\Sigma^{-1})_{uv} = 0$.*

Let $\mathcal{C}^T$ denote the set of all covariance matrices that respect these constraints. Note that it is possible to give a constructive description of this set, simply by writing down the factorization in the directed tree $T$ using linear Gaussian conditional probability distributions. The number of degrees of freedom in such a construction is $2m - 1$ if no constraint is imposed on the variance of the components, and $m - 1$ if the components are constrained to have unit variance.

Finally, we can compute $I^T(x)$ for a Gaussian variable $x$ with covariance matrix $\Sigma$:

$$I^T(\Sigma) \triangleq I^T(x) = I^G(\Sigma) - \sum_{(u,v) \in T} I^G_{uv}(\Sigma) \quad (7)$$

where $I^G(\Sigma) \triangleq -\frac{1}{2} \log \frac{\det \Sigma}{\Sigma_{11} \cdots \Sigma_{mm}}$ is the $m$-fold mutual information between $x_1, \ldots, x_m$ and $I^G_{uv}(\Sigma) \triangleq -\frac{1}{2} \log \frac{\Sigma_{uu}\Sigma_{vv} - \Sigma_{uv}^2}{\Sigma_{uu}\Sigma_{vv}}$ is the pairwise mutual information between $x_u$ and $x_v$. We then have the appealing property that for any positive definite matrix $\Sigma$, $\Sigma \in \mathcal{C}^T$ if and only if $I^T(\Sigma) = 0$. Once the set $\mathcal{C}^T$ is well defined, we can easily solve TCA in the Gaussian case, as the following theorem makes precise:

**Theorem 3** *If $x_1, \ldots, x_m$ are jointly Gaussian with covariance matrix $\Sigma$, the variable $s = Wx$ factorizes in the tree $T$ if and only if there exists a rotation matrix $R$ and $C \in \mathcal{C}^T$ such that $W = C^{1/2}R\Sigma^{-1/2}$.*

The study of the Gaussian case is useful for two reasons. First, we will use Eq. (7) to define the KGV contrast function in Section 5. Second, the Gaussian solution can be exploited to yield a principled reduction of the search space for $W$. Recall that in ICA it is common to decorrelate the data in a pre-processing step; once this is done the matrix $W$ can be constrained to be a rotation matrix. In the TCA model, we cannot require decorrelation of the components (indeed, two components linked by an edge are heavily correlated), but for a given tree $T$, it seems reasonable to constrain $W$ to be such that it is a solution to the Gaussian relaxation of the problem. This cannot be achieved entirely, because if a distribution factorizes in $T$, a Gaussian variable with the same first and second order moments does not necessarily factorize in $T$ (marginal independences are preserved, but conditional independences are not); nevertheless, such a reduction in complexity in the early stages of the search is very helpful to the scalability of our methods to large $m$, as discussed in Section 9. We now turn to the definitions of our two empirical contrast functions for TCA.

## 4　ESTIMATING ENTROPIES USING DENSITY ESTIMATION

Our first approach to approximating the objective function $J(x, W, T)$ is a direct approach, based on approximating the component marginal entropies $H(s_i)$ and joint entropies $H(s_i, s_j)$, $H(s)$ and $H(x)$ via kernel density estimation. The first term in $J(x, W, T)$ can be written $I(s_1, \ldots, s_m) = \sum_i H(s_i) - H(s)$, which can be expanded into $\sum_i H(s_i) - H(x) - \log|\det W|$. Since the joint entropy $H(x)$ is constant we do not need to compute it and thus to estimate $I(s_1, \ldots, s_m)$ we need only estimate one-dimensional entropies $H(s_i)$. We also require estimates of the pairwise mutual informations in the definition of $J(x, W, T)$, which we obtain using two-dimensional entropy estimates. Thus, letting $\hat{H}_i$ and $\hat{H}_{uv}$ denote estimates of the singleton and pairwise entropies, respectively, we define the following contrast function:

$$J^E \triangleq \sum_i \hat{H}_i - \sum_{(u,v) \in T} (\hat{H}_u + \hat{H}_v - \hat{H}_{uv}) - \log|\det W|. \quad (8)$$

Given a set of $N$ training samples $\{x_i\}$ in $\mathbb{R}^d$ and a kernel in $\mathbb{R}^d$, that is, a nonnegative function $K : \mathbb{R}^d \to \mathbb{R}$ that integrates to one, the kernel density estimate with bandwidth $h$ is defined as $\hat{f}(x) = \frac{1}{Nh^d} \sum_{i=1}^N K\left(\frac{x - x_i}{h}\right)$ (Silverman, 1985). In this paper we use Gaussian kernels $K(x) = \frac{1}{(2\pi)^{d/2}} e^{-\|x\|^2/2}$ with $d = 1, 2$.

The entropy of $x$ is estimated by evaluating $\hat{f}(x)$ at points on a regular mesh that spans the support of the data, and then performing numerical integration. The fast Fourier transform can be used to speed up the evaluations, resulting in a complexity which is linear in $N$ and depends on the number $M$ of grid points that are required.

Although automatic methods exist for selecting the bandwidth $h$ (Silverman, 1985), in our experiments we used a constant $h = 0.125$. We also fixed $M = 256$.

Although each density estimate can be obtained reasonably cheaply, we have to perform $O(m^2)$ of these when minimizing over the tree $T$. This can become expensive for large $m$, although, as we show in Section 9, this cost can be reduced by performing optimization on subtrees. In any case, our second contrast function is aimed at handling problems with large $m$. We describe this contrast function in the following section.

## 5 KERNEL GENERALIZED VARIANCE

Our second contrast function is based on the *kernel generalized variance*, an approximation to mutual information introduced by Bach and Jordan (2001). We begin by reviewing the kernel generalized variance, emphasizing a simple intuitive interpretation. For precise definitions and properties, see Bach and Jordan (2001).

### 5.1 APPROXIMATION OF MUTUAL INFORMATION

Let $x = (x_1, \ldots, x_m)^T$ be a random vector with covariance matrix $\Sigma$. The $m$-fold mutual information for an associated Gaussian random vector with same mean and covariance matrix as $x$ is equal to $I^G(\Sigma) = -\frac{1}{2} \log \left( \frac{\det \Sigma}{\Sigma_{11} \cdots \Sigma_{mm}} \right)$. The ratio $\frac{\det \Sigma}{\Sigma_{11} \cdots \Sigma_{mm}}$ is usually referred to as the *generalized variance*.

If $x$ is not Gaussian, $I^G$ is not in general a good approximation to the mutual information between $x_1, \ldots, x_m$. But if we first map each component $x_i$ from $\mathbb{R}$ to a higher-dimensional space $\mathcal{F}$, and then treat the mapped variables as Gaussian in this space, it turns out that we obtain a useful approximation of the mutual information. More precisely, we map each component $x_i$ to $\Phi(x_i) \in \mathcal{F}$ via a map $\Phi$, and define the covariance matrix $\mathcal{K}$ of $\Phi(x) \triangleq (\Phi(x_1), \ldots, \Phi(x_m))^T \in \mathcal{F}^m$ by blocks: $\mathcal{K}_{ij}$ is the covariance between $\Phi(x_i)$ and $\Phi(x_j)$. The size of each of these blocks is the dimension of $\mathcal{F}$. For simplicity we can think of $\mathcal{F}$ as a finite-dimensional space, but the definition can be generalized to any infinite-dimensional Hilbert space. Let $\Phi^G(x) \in \mathcal{F}^m$ be a Gaussian random vector with the same mean and covariance as $\Phi(x)$. The mutual information $I^K(\mathcal{K})$ between $\Phi_1^G(x), \ldots, \Phi_m^G(x)$ is equal to

$$I^K(\mathcal{K}) = -\frac{1}{2} \log \frac{\det \mathcal{K}}{\det \mathcal{K}_{11} \cdots \det \mathcal{K}_{mm}}, \quad (9)$$

where the ratio $\frac{\det \mathcal{K}}{\det \mathcal{K}_{11} \cdots \det \mathcal{K}_{mm}}$ is called the *kernel generalized variance (KGV)*.

We now list the main properties of $I^K$, which we refer to as the *KGV-mutual information*.

**Mercer kernels and Gram matrices.** A *Mercer kernel* on $\mathbb{R}$ is a function $k(x, y)$ from $\mathbb{R}^2$ to $\mathbb{R}$ such that for any set of points $\{x^1, \ldots, x^N\}$ in $\mathbb{R}$, the $N \times N$ matrix $K$, defined by $K_{ij} = k(x_i, x_j)$, is positive semidefinite. The matrix $K$ is usually referred to as the *Gram matrix* of the points $\{x^i\}$. Given a Mercer kernel $k(x, y)$, it possible to find a space $\mathcal{F}$ and a map $\Phi$ from $\mathbb{R}$ to $\mathcal{F}$, such that $k(x, y)$ is the dot product in $\mathcal{F}$ between $\Phi(x)$ and $\Phi(y)$ (see, e.g., Schölkopf and Smola, 2001). The space $\mathcal{F}$ is usually referred to as the *feature space* and the map $\Phi$ as the *feature map*. This allows us, given sample data, to define an estimator of the KGV via the Gram matrices $K_i$ of each component $x_i$. Indeed, using the "kernel trick," we can find a basis of $\mathcal{F}$ where $\mathcal{K}_{ij} = K_i K_j$. Thus, for the remainder of the paper, we assume that $\Phi$ and $\mathcal{F}$ are associated with a Mercer kernel $k(x, y)$.

**Linear time computation.** If $k(x, y)$ is the Gaussian kernel $k(x, y) = \exp(-(x - y)^2 / 2\sigma^2)$, then the estimator based on Gram matrices can be computed in linear time in the number $N$ of samples. In this situation, $\mathcal{F}$ is an infinite-dimensional space of smooth functions. This low complexity is obtained through low-rank approximation of the Gram matrices using incomplete Cholesky decomposition (Bach and Jordan, 2001). We need to perform $m$ such decompositions, where each decomposition is $O(N)$. The worst-case running time complexity is $O(mN + m^3)$, but under a wide range of situations, the Cholesky decompositions are the practical bottlenecks of the evaluation of $I^K$, so that the empirical complexity is $O(mN)$.

**Relation to actual mutual information**. For $m = 2$, when the kernel width $\sigma$ tends to zero, the KGV mutual information tends to a quantity that is an expansion of the actual mutual information around independence (Bach and Jordan, 2001). In addition, for any $m$, $I^K$ is a valid contrast function for ICA, in the sense that it is equal to zero if and only if the variables $x_1, \ldots, x_m$ are pairwise independent.

**Regularization**. For numerical and statistical reasons, the KGV has to be regularized, which amounts to convolving the Gaussian variable $\Phi^G(x)$ by another Gaussian having identity covariance matrix $\kappa I$. This implies that in the approximation of the KGV, we let $\mathcal{K}_{ij} = K_i K_j$ for $i \neq j$, and $\mathcal{K}_{ij} = (K_i + \kappa I)^2$ for $i = j$.

## 5.2 A KGV CONTRAST FUNCTION FOR TCA

Mimicking the definition of the $T$-mutual information in Eq. (4) and the Gaussian version in Eq. (7), we define the KGV contrast function $J^K(x, T)$ for TCA as $J^K(x, T) \triangleq I^K(x) - \sum_{(u,v) \in T} I^K_{uv}(x)$, that is:

$$J^K(x, T) = -\frac{1}{2} \log \frac{\det \mathcal{K}}{\det \mathcal{K}_{11} \cdots \det \mathcal{K}_{mm}} + \frac{1}{2} \sum_{(u,v) \in T} \log \frac{\det \mathcal{K}_{uv,uv}}{\det \mathcal{K}_{uu} \det \mathcal{K}_{vv}}.$$

An important feature of $J^K$ is that it is the $T$-mutual information of $\Phi_1^G(x), \ldots, \Phi_m^G(x)$, which are linked to $x_1, \ldots, x_m$ by the feature maps and the "Gaussianization." It is thus always nonnegative. Note that going from a random vector $y$ to its associated Gaussian $y^G$ is a mapping from distribution to distribution, and not a mapping from each realization of $y$ to a realization of $y^G$. Unfortunately, this mapping preserves marginal independences but not conditional independences, as pointed out in Section 3.2. Consequently, $J^K(x, T)$ does not characterize factorization in $T$; that is, $J^K(x, T)$ might be strictly positive even when $x$ does factorize in $T$. Nonetheless, based on our earlier experience with KGV in the case of ICA (Bach and Jordan, 2001), we expect $J^K(x, T)$ to provide a reasonable approximation to $I^T(x)$. Intuitively, we fit the best tree for the Gaussians in the feature space and hope that it will also be a good tree in the input space.

Numerically, $J^K(x, T)$ behaves particularly nicely, since all of the quantities needed are Gram matrices, and are obtained from the $m$ incomplete Cholesky decompositions. Thus we avoid the $O(m^2)$ complexity.

In our empirical experiments, we used the settings $\sigma = .5$ and $\kappa = .001$ for the free parameters in the KGV, taken from Bach and Jordan (2001). The contrast function that we minimize with respect to $T$ and $W$ is then $J^K(x, W, T) \triangleq J^K(Wx, T)$.

## 6 THE TCA ALGORITHM

We now give a full description of the TCA algorithm. Either of the two contrast functions that we have defined can be used in the algorithm. We generically denote the contrast function as $J(x, W, T)$ in the following description of the algorithm.

**Formulation of the optimization problem**. First, we minimize $J(x, W, T)$ on the space of matrices such that $Wx$ has unit variance components. That is, if $\Sigma$ denotes the covariance matrix of $x$, we constrain the rows of $W\Sigma^{1/2}$ to have unit norm. We also add a penalty term, $J^C(x, W, T) \triangleq -\frac{1}{2} \sum_{(u,v) \in T} \log \left(1 - \text{corr}^2((Wx)_u, (Wx)_v)\right)$ that penalizes marginal correlation along edges of the tree $T$. We thus aim to solve the following optimization problem over $W$ and $T$:

$$\begin{aligned}\text{minimize} \quad & J(x, W, T) + \lambda_C J^C(x, W, T) \\ \text{subject to} \quad & (W\Sigma W^T)_{ii} = 1, \; \forall i \in \{1, \ldots, m\}\end{aligned}$$

where $\lambda_C$ determines how much we penalize the marginal correlations. In all of our experiments we used $\lambda_C = 0.05$.

**Alternating minimization**. We solve the optimization problem by alternatively minimizing with respect to $W$ and $T$. For $W$ fixed, minimizing with respect to $T$ is equivalent to a maximum weight spanning tree problem, while for $T$ fixed, optimization is performed using steepest descent with line search. Note that the empirical gradient can be computed very efficiently, without having to recompute all the Cholesky decompositions (for KGV) or the bivariate densities (for KDE). In our implementation, we performed minimization with respect to $T$ after each line search in the direction of the gradient with respect to $W$. Thus, after each line search we may switch trees.

**Initialization using ICA**. As is the case in ICA, we expect to be faced with local minima. Fortunately, we can obtain a good initialization for TCA using the result of any ICA algorithm. Intuitively, this is helpful in our setting, because ICA is known to find for components that are as "non-Gaussian" as possible (Hyvärinen et al., 2001), and therefore the components that ICA finds should be linear combinations of only a few of the original non-Gaussian components (combinations of large numbers of components are subject to the central limit theorem and should approach Gaussianity). Thus by initializing with an ICA solution, the search for a TCA solution can effectively be limited to a subspace of lower dimension. Additional optimization techniques are described in Section 9.

## 7 DENSITY ESTIMATION

Once the linear transformation $W$ and the tree $T$ are determined, we are faced with a simple density estimation problem. The model with respect to which we carry out this density estimation is a tree, and thus we can work either within the directed graphical model framework or the undirected graphical model framework. We prefer the former because the lack of a normalizing constant implies that the density estimation problem decouples.

We thus have a density of the form $p(s) = p(s_1) \prod_{u=2}^{m} p(s_u | s_{\pi_u})$ to estimate, where $\pi_u$ is the (only) parent of node $u$ in the directed tree $T$. The overall estimation problem reduces to finding $m$ distinct density estimates: a one-dimensional estimate is needed for the root of the tree, and $m-1$ conditional density estimates are needed for the remaining nodes. In this paper we use a Gaussian mixture model for the density at the root, and conditional Gaussian mixture models, also known as "mixtures of experts models" (Jacobs et al., 1991), for the remaining conditional probabilities.

All of these mixture models can be estimated via the expectation-maximization (EM) algorithm. In order to determine the number of mixing components for each model, we use the minimum description length criterion (Rissanen, 1978).

## 8 SIMULATION RESULTS

We have conducted an extensive set of experiments using synthetic data. In a first set of experiments, we focus on the performance of the first stage of the algorithm, i.e., the optimization with respect to $W$ and $T$, when the data actually follow the TCA model. In a second set of experiments, we focus on the density estimation performance, in situations where the TCA model assumptions actually hold, and in situations where they do not.

Table 1: Recovering the tree $T$ and the matrix $W$, for increasing numbers $m$ of components. See text for details on the definitions of $e_W$ and $e_T$.

| $m$ | $e_W$-kgv | $e_W$-kde | $e_T$-kgv | $e_T$-kde |
|---|---|---|---|---|
| 4  | 3.2 | 3.9 | 4.6 % | 2.0 % |
| 6  | 4.7 | 3.1 | 8.1 % | 6.3 % |
| 8  | 6.3 | 4.4 | 9.3 % | 5.7 % |
| 12 | 6.2 | 2.1 | 9.1 % | 3.3 % |
| 16 | 6.8 | 2.3 | 11 %  | 5.1 % |

### 8.1 RECOVERING THE TREE AND THE LINEAR TRANSFORM

In this set of experiments, for various numbers of variables $m$, we generated data from distributions with known density: we selected a tree $T$ at random, and conditional distributions were selected among a given set of mixtures of experts. Then data were generated and rotated using a known random matrix $W$.

To evaluate the results $\hat{W}$ and $\hat{T}$ of the TCA algorithm, we need to use error measures that are invariant with respect to the known invariances of the model, as discussed in Section 3.1. For the demixing matrix $W$, we use a measure commonly used for ICA (Amari et al., 1996), that is invariant by permutation and scaling of rows: we form $A = \hat{W} W^{-1}$ and compute $d = \frac{100}{2m(m-1)} \left[ \sum_{i=1}^{m} \frac{\sum_{j=1}^{m} |a_{ij}|}{\max_j |a_{ij}|} + \sum_{j=1}^{m} \frac{\sum_{i=1}^{m} |a_{ij}|}{\max_i |a_{ij}|} - 2m \right]$. It is always between 0 and 100 and equal to 0 if and only if there is a perfect match between $W$ and $\hat{W}$. However, because of the "leaf mixing" invariance, before computing $d$, we transform $W$ and $\hat{W}$ to equivalent demixing matrices which respect the normalization we choose—marginal decorrelation between the leaf node and its parent. We let $e_W$ denote the final error measure.

For the tree $T$, we use an error measure that is invariant by permutation. We define $e_T = 1 - \frac{s_T - 1}{m - 1}$ where $s_T$ is the size of the largest common connected subtree of $T$ and $\hat{T}$, up to permutation of the nodes, i.e., $s_T$ is the maximum integer $s$ for which there exists a connected subtree of $T$ of size $s$ that can be mapped to a connected subtree of $\hat{T}$. Intuitively, $e_T$ is the fraction of wrong edges. We report results in Table 1. We used 1000 samples for $m = 4$, 2000 samples for $m = 6, 8$ and 4000 samples for $m = 12, 16$. All results are averaged over 20 replications. Our algorithms manage to recover $W$ and $T$ very accurately. An interesting fact that is not apparent in the table is that our results are quite insensitive to the "density" of the tree that was used to generate the data: bushy trees yield roughly the same performance as sparse trees.

Table 2: Density estimation for increasing number of components $m$ and treewidth $\tau$ of the generating model (all results are averaged over 20 replications).

| $m,\tau$ | GAU | IND | CL | ICA | GMM | T-kgv | T-kde |
|---|---|---|---|---|---|---|---|
| **4, 1** | **1.3** | **1.3** | **0.93** | **1.1** | **0.44** | **0.20** | **0.25** |
| 4, 2 | 1.1 | 1.0 | 0.58 | 0.77 | 0.22 | 0.33 | 0.26 |
| **6, 1** | **2.4** | **2.4** | **2.0** | **1.9** | **0.90** | **0.45** | **0.34** |
| 6, 2 | 1.8 | 2.1 | 1.5 | 1.5 | 0.60 | 0.43 | 0.28 |
| 6, 3 | 1.3 | 1.5 | 1.1 | 1.0 | 0.59 | 0.47 | 0.39 |
| 6, 4 | 1.0 | 1.1 | 0.89 | 0.81 | 0.43 | 0.66 | 0.33 |
| **8, 1** | **3.3** | **3.4** | **3.0** | **2.7** | **1.9** | **0.99** | **0.70** |
| 8, 2 | 2.6 | 3.1 | 2.5 | 2.2 | 1.0 | 0.87 | 0.38 |
| 8, 3 | 2.1 | 2.3 | 2.0 | 1.7 | 0.99 | 0.88 | 0.33 |
| 8, 4 | 2.0 | 2.2 | 1.9 | 1.6 | 1.0 | 0.91 | 0.38 |

## 8.2 DENSITY ESTIMATION

Here we focus on density estimation, comparing the following models: Gaussian (GAU), Gaussian mixture (GMM), independent Gaussian mixtures (IND), Chow-Liu with Gaussian mixtures (CL), ICA using marginal Gaussian mixtures (ICA), and TCA using Gaussian mixtures (T-kde, T-kgv).

We generated data as follows: we designed a set of graphical models with given treewidth $\tau$ between 1 (trees) and 4 (maximal cliques of size 5). Then data were generated using one of these models and rotated using a known random matrix $W$. We report results in Table 2, where performance is measured as the average loglikelihood of a held-out test set, minus the same average under the (known) generating model.

When the treewidth $\tau$ is equal to 1 (lines in bold in Table 2), the data exactly follow the TCA model and it is no surprise that our TCA algorithm outperforms the other models. However, when $\tau$ is greater than one, the TCA model assumptions do not hold, but our models still exhibit good performance, with a slight advantage for KDE over KGV. Note that when the generating model becomes too connected (e.g., $m = 6, \tau = 4$), the performance of the TCA models starts to degrade, which simply illustrates the fact that in those conditions the tree approximation is too loose.

## 9 SCALING ISSUES

In the experiments reported in this paper, we have limited ourselves to problems in which the number of components $m$ is less than 16. While the algorithms presented in Section 6 can be directly applied to large $m$, for problems in which $m$ is significantly larger, additional numerical techniques are needed, for two reasons. The first one is running time complexity. Indeed, the contrast function based on KGV scales as $O(mN)$ but the one based on KDE scales as $O(m^2 N)$. Second, with increasing number of sources, both ICA and TCA contrast functions tend to have multiple local minima. In Section 6, we already mentioned the use of the result of ICA as an initialization to TCA. We now present additional optimization techniques aimed at dealing with large-scale problems.

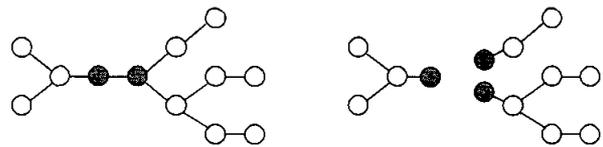

Figure 1: (Left) original tree $T$ and subtree $V$ (shaded). (Right) subtrees $U_i$ (non shaded) with their neighbor in $T$ (shaded)

**Optimizing subtrees.** Gradient descent can be performed sequentially on limited subspaces of the space of the matrix $W$. Indeed, given current estimates for $T$ and $W$, we can perform optimization over a subset of the rows of $W$ whose indices span a connected subtree $U$ of $T$. In this case, the overall contrast function can be approximated by the contrast function for the subtree that contains $U$ and its neighbors in $T$.

To select the subtrees $U$, we use the following procedure: we generate all the subtrees $V$ of small sizes (less than 4) and we measure how well $V$ "separates the graph"; that is, if we let $U_1, \ldots, U_p$ denote the connected components of the complement of $V$ in $T$ (see Figure 1), and if $s = Wx$ are the estimated sources, we measure the conditional independence of $s_{U_1}, \ldots, s_{U_p}$ given $s_V$. The KGV provides such a measure, by simply computing $J(V) = I(s_V, s_{U_1}, \ldots, s_{U_p}) - \sum_{i=1}^{p} I(s_V, s_{U_i})$, where the mutual informations are estimated using the Gaussian variables in feature space: once the Cholesky decompositions of each component $s_i$ are performed and cached, computing all these scores only involves determinants of small matrices, and thus many subtrees $V$ can be scored. The subtrees $V$ with small score $J(V)$ do not need to be improved, and the subtrees that are selected for further optimization are the connected components $U_1, \ldots, U_p$ corresponding to those subtrees $V$.

**Cumulant-based contrast functions.** A contrast function based on cumulants is easily derived from Eq. (8), using Gram-Charlier expansions to compute one-dimensional and two-dimensional entropies, as discussed by Akaho et al. (1999) and Cardoso (1999). Since this contrast function only involves up to fourth order cumulants, it is numerically efficient and can be used to rapidly find an approximate solution which can serve as an initialization for the slower but more accurate contrast functions based on KGV or KDE.

**Covariance constraint.** We can constrain the $W$ matrix to yield a solution in the Gaussian case, as detailed in Section 3.2. We can optimize over matrices that belong to $\mathcal{C}^T$ and thus reduce the dimension of the search space from $m(m-1)$ to $m(m-1)/2 + (m-1)$.

## 10 CONCLUSION

We have presented a model that extends the classical ICA model, by allowing tree-structured dependence among the components. The tree $T$ and the demixing matrix $W$ are determined by minimizing contrast functions within a semiparametric estimation framework. Once $W$ and $T$ are found, the remaining densities are readily estimated. Our two-stage approach is to be contrasted with an alternative one-stage approach in which one would define a model using $W$, $T$ and Gaussian-mixture conditional distributions, and perform maximum likelihood using EM—such an approach could be viewed as an extension of the independent factor analysis model (Attias, 1999) to the tree setting. By separating density estimation from the search for $W$ and $T$, however, we are able to exploit the reduction of our parameter estimation problem to bivariate density estimation. Bivariate density estimation is a well-studied problem, and it is possible to exploit any of a number of parametric or nonparametric techniques. These techniques are computationally efficient, and good methods are available for controlling smoothness. We also are able to exploit the KGV technique within the two-phase approach, an alternative that does not rely on density estimation.

Finally, and perhaps most significantly, our approach does not lead to intractable inference problems that require sampling or variational methods, as would be necessary within a tree-based generalization of the independent factor analysis approach.

Clearly, there are a number of potential generalizations of the methods discussed in this paper. In general, we believe that TCA and ICA provide appealing examples of the utility of applying an adaptive transformation to data before fitting a model, and thereby extending the range of problems to which graphical models can be usefully applied. Moreover, kernel generalized variances provide a fast and flexible way of computing model scores, not only for continuous variables but potentially also for discrete variables and discrete structures, such as strings and trees (Lodhi et al., 2001). Finally, although we have limited ourselves to a generalization of ICA that allows tree-structured dependency among the latent variables, it is clearly of interest to make use of the general graphical model toolbox and consider broader classes of dependency.

### Acknowledgments

We would like to acknowledge support from NSF grant IIS-9988642 and ONR MURI N00014-00-1-0637.